\documentclass{article}
\pdfoutput=1

\PassOptionsToPackage{numbers, compress}{natbib}
\usepackage[preprint]{neurips_2020}

\usepackage[utf8]{inputenc} % allow utf-8 input
\usepackage[T1]{fontenc}    % use 8-bit T1 fonts
\usepackage{hyperref}       % hyperlinks
\usepackage{url}            % simple URL typesetting
\usepackage{booktabs}       % professional-quality tables
\usepackage{amsfonts}       % blackboard math symbols
\usepackage{nicefrac}       % compact symbols for 1/2, etc.
\usepackage{microtype}      % microtypography
\usepackage[pdftex]{graphicx}

\usepackage{amsmath}

\title{Training Deep Spiking Neural Networks}

\author{%
Eimantas Ledinauskas \hskip2em 
Julius Ruseckas \hskip2em
Alfonsas Jur\v{s}\.enas\\
Baltic Institute of Advanced Technology (BPTI)\\
Pilies 16-8, LT-01403, Vilnius, Lithuania\\
\texttt{\{eimantas.ledinauskas, julius.ruseckas, alfonsas.jursenas\}@bpti.eu}\\
\AND
Giedrius Bura\v{c}as\\
SRI International, USA\\
\texttt{giedrius.burachas@sri.com}
}

\begin{document}

\maketitle

\begin{abstract}
Computation using brain-inspired spiking neural networks (SNNs) with
neuromorphic hardware may offer orders of magnitude higher energy efficiency
compared to the current analog neural networks (ANNs). Unfortunately, training
SNNs with the same number of layers as state of the art ANNs remains a
challenge. To our knowledge the only method which is successful in this regard
is supervised training of ANN and then converting it to SNN. In this work we
directly train deep SNNs using backpropagation with surrogate gradient and find
that due to implicitly recurrent nature of feed forward SNN's the exploding or
vanishing gradient problem severely hinders their training. We show that this
problem can be solved by tuning the surrogate gradient function. We also
propose using batch normalization from ANN literature on input currents of SNN
neurons. Using these improvements we show that is is possible to train SNN with
ResNet50 architecture on CIFAR100 and Imagenette object recognition datasets.
The trained SNN falls behind in accuracy compared to analogous ANN but requires
several orders of magnitude less inference time steps (as low as 10) to reach
good accuracy compared to SNNs obtained by conversion from ANN which require on
the order of 1000 time steps.
\end{abstract}

\section{Introduction and related work}
Spiking neural networks (SNNs) are the brain inspired artificial neural network
models that (if implemented in neuromorphic hardware) offer better energy
efficiency due to sparse event-driven and asynchronous information processing
\cite{sorbaro2019optimizing}. It was shown that SNNs computation capacity is
theoretically at least not smaller than analog neural networks (ANN's)
\cite{maass2004computational}.  There are two important differences of SNNs
with respect to analog neural networks (ANN): the activation function is not
differentiable and SNNs are inherently recurrent due to accumulation of
membrane potential.

Recently it was demonstrated that deep convolutional ANN weights could be
transferred to SNN with only a minor loss in visual recognition accuracy
\cite{Diehl2015, sengupta2019going}. SNNs obtained by conversion are
constrained to only use rate encoding and because of that their expressive
capacity may be reduced in comparison to SNNs with more complex temporal
encoding capabilities.  Other drawback of such conversion is that one needs to
use on the order of 1000 forward propagation time steps (during inference
procedure) for SNN to reach a similar image classification accuracy as ANN
counterpart.  This drawback severely limits the computation speed and energy
efficiency benefit of SNNs.  Many inference time steps may be required because
the conversion procedure is designed to use rate encoding and large numbers of
spikes are required in order to reduce the uncertainty of spiking frequency
values.  In addition, ANN architectures are constrained \cite{Diehl2015,
sengupta2019going} (e.g.\ batchnormalization cannot be used) before the
conversion to SNN, this limits ANN performance and upper bound of SNN
performance; When using conversion one can not apply online learning methods,
therefore it is unusable for continual learning, real time adaptation.  All of
mentioned limitations of conversion motivates the search for a way to train
SNNs directly instead of only converting ANN parameters to SNN.

Training SNNs remains a difficult task in comparison to training of ANN's. The
primary reason for this difficulty is the discontinuous activation function
which prevents direct usage of gradient-based optimization.  Despite the
non-differentiable nature of SNN activations, various workarounds enabling the
usage of backpropagation algorithm can be found in the scientific literature.
    
First approach is to constrain the total number of spikes per neuron during
inference \cite{bohte2002error, mostafa2017supervised}, the timing of the
spikes can be differentiated enabling backpropagation. However, the constraint
of certain number of firings per neuron might significantly limit the
expressive capacity of the network.  Second approach is bio-inspired learning
algorithms like spike-timing dependent plasticity for unsupervised learning
\cite{kheradpisheh2018stdp} and reward modulation for supervised learning
\cite{bing2019supervised}. Yet these methods remain not effective enough to
train deep networks with more than a few layers and even with only a single
hidden layer fail to reach the classification accuracy of analogous ANN's
trained with backpropagation algorithm.  Third approach is to approximate the
discontinuous spike activation function as a smooth one during training. This
enables using backpropagation algorithm directly with a cost that the network
must use different activation functions during training and inference
\cite{Neftci2019}.  Fourth approach exploits the observation that the absolute
accuracy of the gradients is not necessary for successful training and uses
surrogate gradients of spike activation functions in place of the real ones
enabling the direct training of true SNNs with backpropagation
\cite{Neftci2019}.

One of the first uses of surrogate derivatives have been described in
\cite{Bohte2011}. Different forms of surrogate gradient have been proposed:
piece-wise linear
$
\beta\max\{0,1-|U-\vartheta|\}
$ \cite{Bohte2011,Esser2016,Bellec2018},
derivative of a fast sigmoid
$
\big[1+|\gamma(U-\vartheta)|\big]^{-2}
$ \cite{Zenke2018},
exponential function
$
\beta \exp\{-\gamma|U-\vartheta|\}
$ \cite{Shrestha2018},
rectangular
$
\beta\mathop{\mathrm{sign}}\{\gamma - |U-\vartheta|\}
$ \cite{Wu2019}. As the experience shows, the specific functional form of surrogate gradient is not important.

To our knowledge, from all of the training approaches described in the previous
paragraphs the methods based on surrogate gradients have achieved the best
results. For example, in \cite{Wu2019} and \cite{Lee2020} the authors managed
to almost reach the accuracy of converted SNN's on CIFAR10 with ResNet11
architecture. However, until now, successful training of very deep
architectures like ResNet50 has not been demonstrated.  Up to this day the best
performance on complex datasets like ImageNet or CIFAR100 with deep SNN
architectures (e.g. ResNet50) has been achieved by supervised training ANN and
then converting it to SNN \cite{sengupta2019going, Lee2020}.

In this article we explore the possibility to directly train very deep SNNs. To
achieve that we use surrogate gradients as proposed in \cite{Neftci2019}. We
found that with increasing number of neuron layers the exploding or vanishing
gradients quickly become a problem severely hindering the training. This might
be explained by the fact that SNN could be understood as an RNN due to
accumulation of membrane potential of a neuron (see
figure~\ref{fig:snn_diagram}). When backpropagation algorithm is used, SNN
inherits vanishing and exploding gradient problems of RNN's due to a deep
unrolled computational graph. We found that this problem can be solved by
tuning the surrogate gradient function. We also propose using batch
normalization on the input currents of the neurons. These simple improvements
let us directly train very deep SNNs (e.g. ResNet50) on complex datasets like
CIFAR100 \cite{krizhevsky2009learning} or Imagenette \cite{imagenette}. To our
knowledge, such deep SNNs have never been directly trained successfully before
and could only be obtained by conversion from ANN. By training SNNs directly we
are unable to reach the accuracy of analogous ANN's, but we find that our
networks require orders of magnitude less inference time steps (as low as 10)
to already reach good performance. This might be very important in order to
minimize the inference time and energy costs in the real world applications.

\section{Methods}

\subsection{Architecture of spiking neural networks}

\begin{figure}[tbp]
  \centering
  \includegraphics[width=1.0\linewidth]{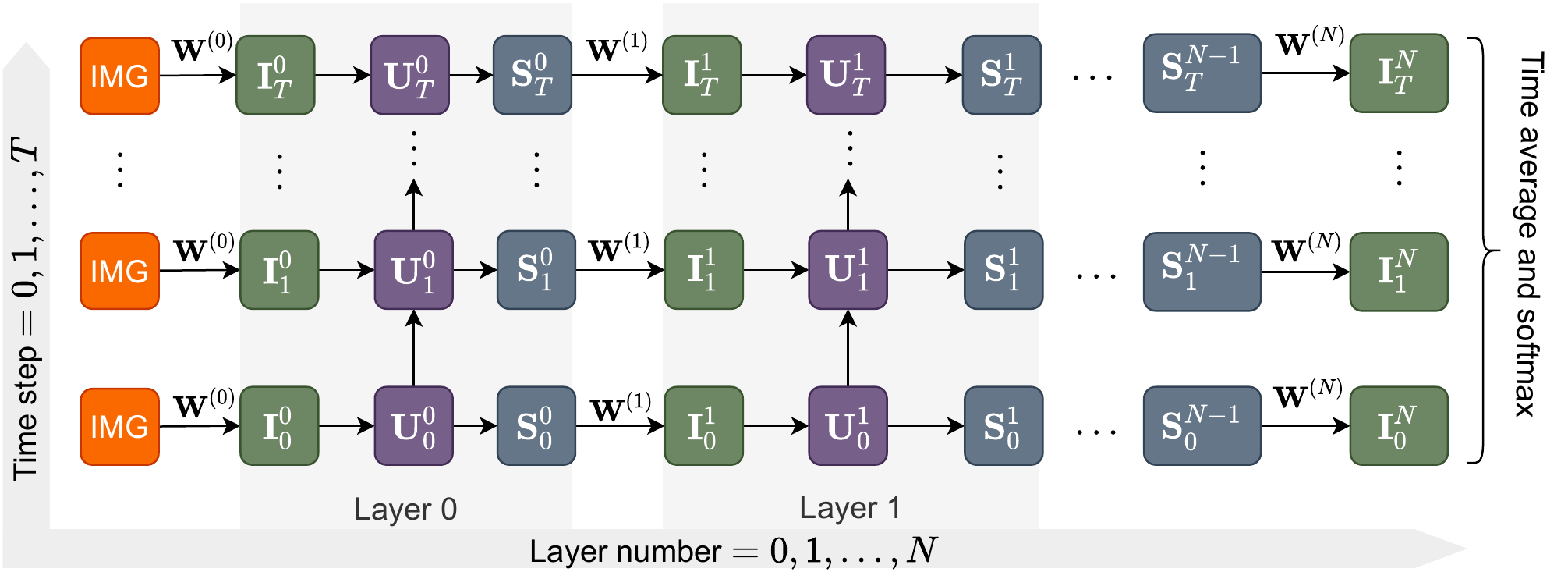}
  \caption{ SNN diagram unrolled in time, here $\mathbf{W}^{(n)}$ denotes the
  weight matrix, $\mathbf{I}_t^n$ - electric currents, $\mathbf{U}_t^n$ -
  synaptic potentials, $\mathbf{S}_t^n$ - spikes, for layer $n$ and time step
  $t$. }
  \label{fig:snn_diagram}
\end{figure}

\begin{figure}[tbp]
  \centering
  \includegraphics[width=\linewidth]{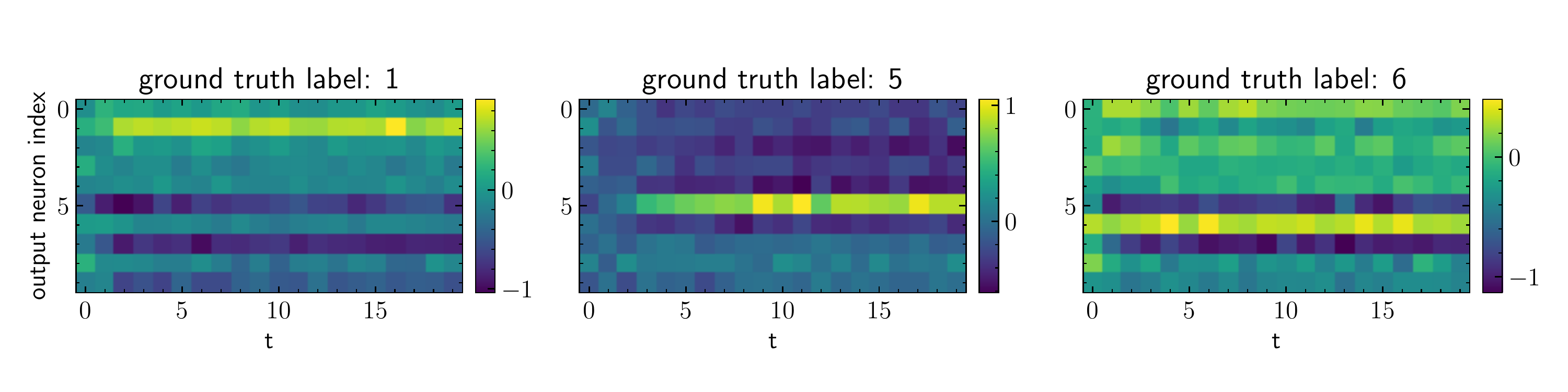}
  \caption{Examples of output currents (in arbitrary units) from the last layer
  neurons of the network trained on CIFAR10 dataset. Vertical axis denotes the
  output neuron index and horizontal axis denotes time step number. Ground truth
  labels of the given examples are written above the images. }
  \label{fig:output}
\end{figure}

In this work we focus on computer vision domain and specifically on image
recognition.  We take common architectures of convolutional neural networks and
replace the nonlinearities by spiking neurons. As is suggested in
\cite{Rueckauer2017}, the analog input image is encoded into a spike train at
the first hidden layer by injecting a constant input current into a spiking
neuron. At the output of the network analog current from the neurons of the
last layer are summed (as in a spiking neuron with very large threshold) and
treated as class logits \cite{Rueckauer2017, Lee2020}. For numerical simulation
we discretize time. In this way SNNs constitute a special case of RNNs with the
time-unrolled computational graph shown in figure~\ref{fig:snn_diagram}. The
typical output of the last layer of an SNN is shown in figure~\ref{fig:output}.

\subsection{Integrate and fire neuron model}
\label{sec:IF_neurons}

In order to minimize the number of possible hyperparameters, in this article we
use a simple integrate and fire (IF) neuron model. Such a model is sufficient
for time-independent classification tasks. For more complicated, time-dependent
tasks more complex neuron models, at least leaky integrate and fire (LIF),
might be needed.

The dynamics of the membrane potential $U(t)$ of IF neuron are described
by the equations \cite{Gerstner2014}
\begin{equation}
C\frac{d}{dt}U_{j}(t)=I_{j}(t)\,,\qquad I_{j}(t)=\sum_{i}W_{ji}S_{i}(t)\,,\label{eq:if-neuron}
\end{equation}
where $C$ is membrane capacitance and $I_{j}(t)$ is the input synaptic current.
The input current is a weighted sum of pre-neuron spikes $S_{i}(t)$, with the
connecting weights being $W_{ji}$. When the membrane potential $U$ reaches the
threshold value $\vartheta$, the neuron generates an output spike and the
membrane potential $U$ is reduced to the rest potential $U_{\mathrm{rest}}$.
We modify equation (\ref{eq:if-neuron}) to be evaluated in discrete time steps
replacing the differential equation by the difference equation:
\begin{equation}
u_{j}^{t}=u_{j}^{t-1}+\sum_{i}w_{ji}s_{i}^{t-1}\,.\label{eq:if-discrete}
\end{equation}
The output spike is $s_{j}^{t}=1$ if $u_{j}^{t-1}>\vartheta$ and $s_{j}^{t}=0$
otherwise.  Once the spike $s_{j}^{t}$ is generated, the membrane potential is
reset.  In this article we consider two types of reset: reset to zero
$u_{j}^{t+1}=u_{j}^{t}(1-s_{j}^{t})$ (hard reset) \cite{Diehl2015} and reset by
subtraction $u_{j}^{t+1}=u_{j}^{t}-\vartheta s_{j}^{t}$ (soft reset)
\cite{Cassidy2013,Diehl2016}.  It has been argued \cite{Rueckauer2017,Han2020}
that soft reset reduces the information loss and improves ANN-SNN conversion.
We found that training shallow (several layers) models with soft reset allows
us to obtain better classification accuracy, compared to models with hard
reset. However in deep models hard reset performs better. We hypothesize that
the advantage of hard reset for deeper models might be related to the
accumulation of inaccuracies due to surrogate gradient (see section
\ref{sec:discussion}) because after a hard reset the time history of the
membrane potential is forgotten.

\subsection{Surrogate gradient for training a single neuron}

Let us consider at first a single neuron and investigate how to train
the neuron for emitting a given spike train $S_{\mathrm{gt}}(t)$
for a given stimulus. Employing a physical analogy we propose to use
as a loss function the energy of the error that we define as follows:
the energy is a time integral of power which is equal to the error
current $S_{\mathrm{err}}(t)=S(t)-S_{\mathrm{gt}}(t)$ multiplied
by the membrane potential $U(t)$. Thus we consider the loss
\begin{equation}
\mathcal{L}=\int_{0}^{T}S_{\mathrm{err}}(t)U(t)\,dt
=\int_{0}^{T}\bigl(S(t)-S_{\mathrm{gt}}(t)\bigr)U(t)\,dt\label{eq:loss-1}
\end{equation}
Here $T$ is the total duration of the spike train. Such a loss does not have a
problem with the non-differentiability of the spiking nonlinearity because the
gradient of the output $S(t)$ is zero almost everywhere and can be neglected.
The gradient of this loss is similar to the gradient in SuperSpike
\cite{Zenke2018}, with the function $\sigma(U)=U$.  One can see that the loss
(\ref{eq:loss-1}) has some desirable properties: the loss decreases with
increasing membrane potential when the output pulse is missing
($S-S_{\mathrm{gt}}<0$) thus causing the appearance of the output pulse. If the
output pulse should not be present ($S-S_{\mathrm{gt}}>0$), the loss decreases
with decreasing membrane potential. The drawback of such a loss is that it can
acquire both positive and negative values, and values close to zero can be
obtained when membrane potential is close to zero even if output pulses differ
from the desired ones. However, the optimal values of the weights still
correspond to the gradient of the loss being zero.

In analogy with the van Rossum distance we can define a more general
loss
\begin{equation}
\mathcal{L}=\int_{0}^{T}\bigl[(\alpha*S)(t)-(\alpha*S_{\mathrm{gt}})(t)\bigr](\alpha*U)(t)\,dt\,,
\label{eq:loss-2}
\end{equation}
where
\begin{equation}
(\alpha*X)(t)\equiv\int_{0}^{t}\alpha(t-t')X(t')\,dt'
\end{equation}
denotes a convolution.  A single neuron can be used also in classification
tasks, for example to determine if the number of pulses is larger than a given
threshold. In analogy with equation (\ref{eq:loss-1}), we take the convolution
kernel $\alpha(t)=1$ for whole duration $T$ and propose the following loss for
classification tasks:
\begin{equation}
\mathcal{L}=[\Theta(Y)-Y_{\mathrm{gt}}]\int_{0}^{T}U(t)\,dt\,,\qquad
Y=\int_{0}^{T}S(t)\,dt\,.
\label{eq:loss-3}
\end{equation}
Here $\Theta$ is a Heaviside step function, $Y_{\mathrm{gt}}=0,1$
is a ground-truth label.

The gradient of the loss function (\ref{eq:loss-1}) can be obtained in the
following way: we can consider the functions
\[
\mathcal{L}^{\prime}=\frac{1}{2}\int_{0}^{T}
\bigl[S(t)-S_{\mathrm{gt}}(t)\bigr]^{2}\,dt
\]
instead of (\ref{eq:loss-1})
and
\[
\mathcal{L}^{\prime}=\frac{1}{2}[\Theta(Y)-Y_{\mathrm{gt}}]^{2}
\]
instead of (\ref{eq:loss-3})
together with a replacement rule
\begin{equation}
\nabla S\rightarrow\nabla U\,.\label{eq:sur-grad-1}
\end{equation}
That is, the gradient of a non-differentiable spike is replaced by
another, surrogate, gradient \cite{Neftci2019}.

\subsection{Surrogate gradient for multi-layer spiking neural network}

A way of training SNN with backpropagation is to replace the gradient of a
non-differentiable spike by another, surrogate, gradient \cite{Neftci2019}.
The simplest replacement is just $\nabla S\rightarrow\nabla U$.  However, such
simple surrogate gradient causes problems when applied to multi-layer spiking
neural network. In the situation when the potential of a neuron in a hidden
layer is close to zero, the application of gradient descent with the surrogate
gradient causes the potential to increase or decrease with almost equal
probabilities, and the potential remains close to zero for a long time. In
order to avoid this, we modify the replacement rule as
\begin{equation}
\nabla S\rightarrow f(U)\nabla U\,,\label{eq:sur-grad-2}
\end{equation}
with function $f(U)$ being smaller when the potential is close to zero:
$f(0)<f(\vartheta)$. Numerical experiments show, that the actual form of the
function $f(U)$ is not important. In this article we take
\begin{equation}
f(U)=\beta\big\{1+[\gamma(U-\vartheta)]\big\}^{-2}\,.\label{eq:sur-grad-f}
\end{equation}
Here $\beta$ (we use $\beta=1$) is a hyper-parameter determining the size of
the surrogate gradient and $\gamma$ determines the width of the gradient. As
the discussion above shows, the value $\gamma=0$ is not suitable for deeper
SNNs. On the other hand, very large values of $\gamma$ cause the surrogate
gradient to be almost always very small, except in the rare situations when the
potential $U$ is close to the threshold $\vartheta$. Thus one should expect
that there exist an optimal, intermediate value of $\gamma$.

Note, that in spiking neurons with hard reset we take the gradient of spikes
$s_j^t$ to be zero. In this way the previous values of the membrane potential
$u_j$ do not influence the gradient of the potential after the reset, the
history of the potential is forgotten after reset.

\subsection{Tuning of surrogate gradient}
\label{sec:tuning_gamma}

\begin{figure}[tbp]
  \centering
  \includegraphics[width=0.6\linewidth]{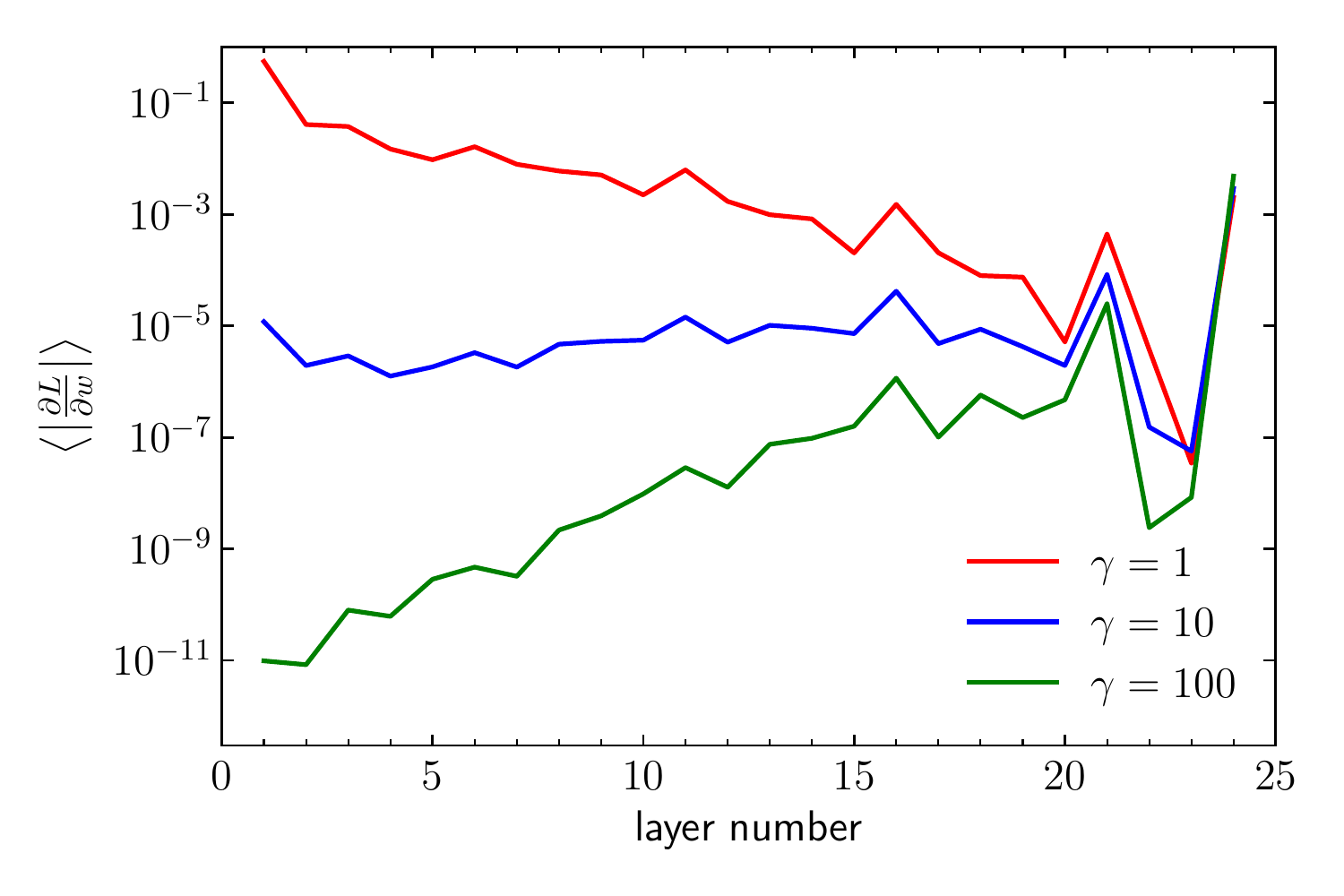}
  \caption{Average magnitude of loss gradient dependence on layer number with
  different values of surrogate gradient parameter $\gamma$. Gradient explodes
if $\gamma$ is too small and vanishes if it is too large. ResNet18 architecture
was used in this picture.}
  \label{fig:tuning}
\end{figure}

Using a surrogate gradient, the gradient of the loss with respect to the
membrane potential $u_{l}$ in the neuron of $l$-th layer is
\begin{equation}
\delta u_{l}=f(u_{l})\delta s_{l}
\end{equation}
where $\delta s_{l}$ is the gradient of the loss with respect to the output
spikes of the $l$-th layer. Similarly as in \cite{He2015}, we assume that the
elements of the gradient $\delta s_{l}$ are mutually independent and share the
same distribution. In addition, we assume that $\delta s_{l}$ and $u_{l}$ are
independent of each other and the gradients have zero means, $\langle\delta
u_{l}\rangle=0$ and $\langle\delta s_{l}\rangle=0$. In this case the variances
of the gradients obey
\begin{equation}
\mathrm{var}[\delta u_{l}]=\langle f(u_{l})^{2}\rangle\mathrm{var}[\delta s_{l}]\,.
\label{eq:variance1}
\end{equation}
If the probability distribution function of $u$ is $P(u)$ with the width $\sigma_{u}$ then
\begin{equation}
\langle f(u)^{2}\rangle=\int_{-\infty}^{+\infty}f(u)^{2}P(u)du
\sim\frac{\beta}{\gamma\sigma_{u}}\,.
\end{equation}
Furthermore, from equation (\ref{eq:if-discrete}) it follows
that the gradient of the loss with respect to the output spikes of
the $l-1$-th layer $\delta s_{l-1}$ is related to the gradient of
the potential of the $l$-th layer $\delta u_{l}$. The variance of
the gradient $\delta s_{l-1}$ is proportional to
\begin{equation}
\mathrm{var}[\delta s_{l-1}]\sim\mathrm{var}[w_{l}]\mathrm{var}[\delta u_{l}]
\sim\mathrm{var}[w_{l}]\langle f(u_{l})^{2}\rangle\mathrm{var}[\delta s_{l}]\,.
\end{equation}
We see that during the backpropagation the variance of the gradient is
multiplied by $\langle f(u)^{2}\rangle$ for each layer. Depending on the value
of $\langle f(u)^{2}\rangle$ this can cause exploding or vanishing gradients in
the deeper neural networks. To avoid this problem we propose to fine-tune the
value of $\langle f(u)^{2}\rangle$. This can be achieved in three ways: i) by
changing the height of the surrogate gradient $\beta$, ii) by changing the
width of the surrogate gradient $\gamma$, iii) by changing the initialization
of the model parameters and thus the resulting probability distribution
function of the neuron potential. In this article we concentrate our attention
on the fine-tuning of the value of the parameter $\gamma$.

An example of the behavior of the gradient for different values of $\gamma$ is
shown in figure~\ref{fig:tuning}. Here we used SNN with ResNet18 architecture.
One can see that for small values of $\gamma=1$ the gradient explodes, whereas
for large values $\gamma=100$ the gradient vanishes. Because of this monotonic
relation the optimal $\gamma$ can be found fast by using bisection search.

\subsection{Batch normalization for SNNs}
\label{sec:batchnorm}

\begin{figure}[tbp]
  \centering
  \includegraphics[width=0.6\linewidth]{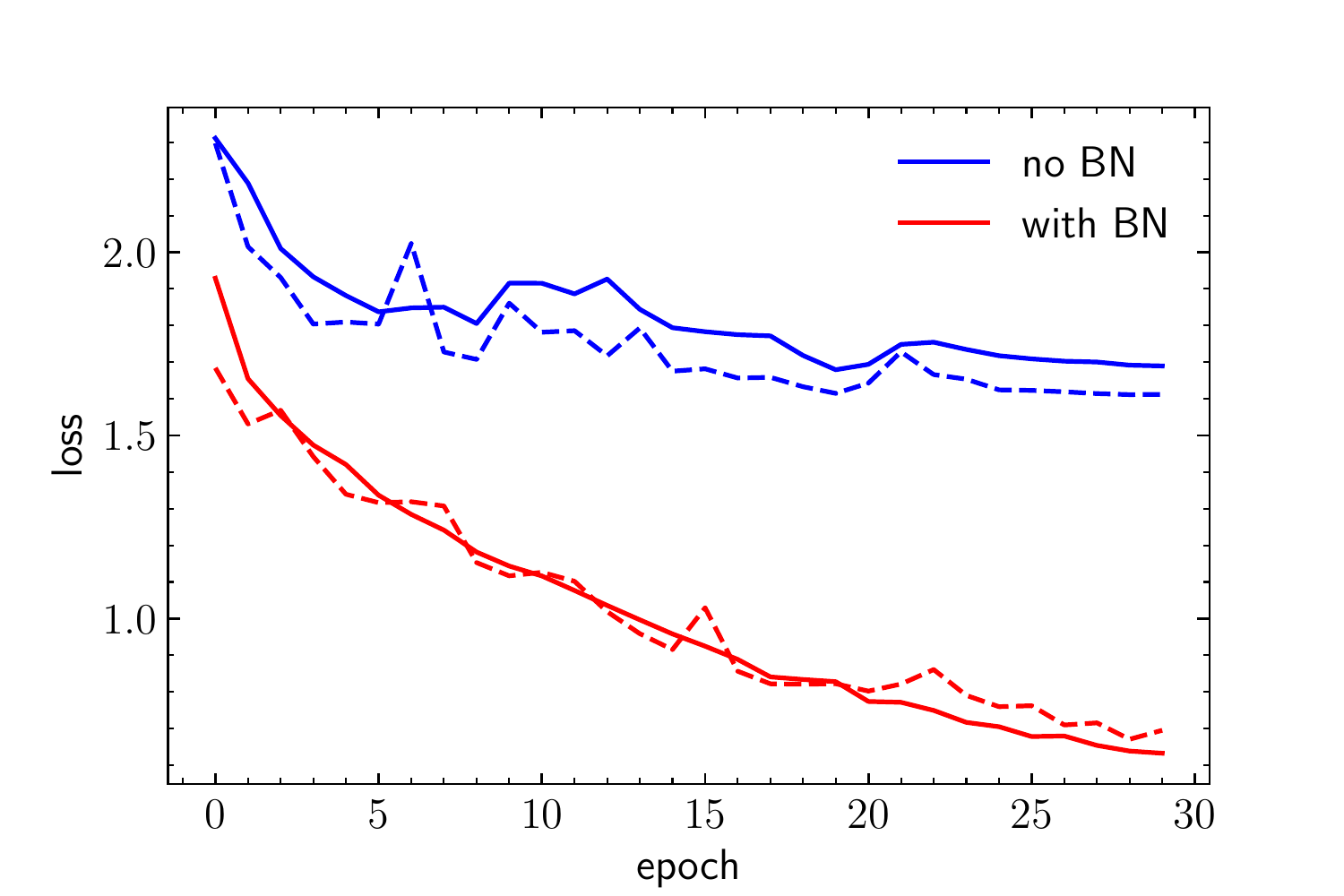}
  \caption{Comparison of typical learning curves for ResNet18 SNN on CIFAR10
  dataset trained with (red) and without (blue) batch normalization. The solid
  and dashed lines denote the training and validation losses respectively.}
  \label{fig:bn_vs_no_bn}
\end{figure}

Batch normalization layer \cite{ioffe2015batch} is a widely used architectural
element in ANN literature. It stabilizes the distribution of inputs to a
certain network layer by setting their mean and variance to some learnable
values. This makes the training of a neural network less sensitive to
hyperparameters and to some extent prevents exploding or vanishing gradients
\cite{ioffe2015batch}. As was shown in \cite{santurkar2018does}, batch
normalization also increases the smoothness of the loss landscape and thus
results in more predictive and well-behaved gradients.

For recurrent neural networks batch normalization has been applied both to the
connections from layer to layer \cite{bn_in_rnn_2016} and to the recurrent ones
\cite{bn_in_rnn2_2016}. Similarly to \cite{bn_in_rnn_2016}, in this work we
apply batch normalization in SNNs on the input currents to every neuron layer
(horizontal connections in figure~\ref{fig:snn_diagram}). In effect, this
allows for the spiking neurons to learn their firing threshold during training
as scaling the input currents is equivalent to changing the threshold. Also
this allows for non-zero bias currents which, depending on the sign of their
value, are equivalent to the leaking or passive accumulation of the potential.
We find that similarly to ANN's the batch normalization in SNNs improves the
rate of convergence and decreases the sensitivity to the hyperparameters.

The examples of typical learning curves for ResNet18 SNN's with and without
batch normalization can be seen in figure~\ref{fig:bn_vs_no_bn}. Clearly the
training converges much faster with batch normalization. We noticed that this
disparity increases even more with increasing network depth. Here in both cases
the networks were trained with ADAMW optimizer \cite{loshchilov2019decoupled}
and fixed learning rate $5 \cdot 10^{-4}$. The network without batch
normalization additionally requires initial normalization of spiking thresholds
because otherwise the spiking activity of the neurons could be too low or too
high. In both cases the training is very slow and might not even happen at all.
To normalize the thresholds we follow the method used in ANN to SNN conversion
\cite{sengupta2019going} and set the threshold of the neuron layer to the
maximum value of the input current to this layer throughout the whole training
dataset and all time steps.

\section{Results}

\subsection{Training methodology}

In this section we describe the general details about training which apply to
all of the models presented in this paper unless stated otherwise.

For optimization we used ADAMW \cite{loshchilov2019decoupled} algorithm with
weight decay parameter 0.01. The learning rate was changed according to
one-cycle learning rate schedule \cite{smith2018disciplined}. The maximum
learning rates for the schedule were chosen by using learning rate range test
as proposed in \cite{smith2018disciplined} for every network separately. We
also used batchnormalization layers between convolution (or fully-connected)
and activation layers for both ANNs and SNNs (see section \ref{sec:batchnorm}).
During SNN training we use 10 time steps for their simulation (justification
for this can be found in \ref{sec:time_steps}). Also for SNNs we fine-tune
gamma so that the average magnitude of the gradient of the layers in the first
half of the network would be as close as possible to the second half (see
\ref{sec:tuning_gamma}). In the cases of CIFAR10 and CIFAR100 we used random
horizontal flip, random crop and color jitter augmentations and in the case of
MNIST we did not use any.

\subsection{Performance comparison with the other works}

In this section we compare the performance achieved by using our method to the
state of the art results found in the literature. Table \ref{tab:mnist_acc}
shows the comparison of classification accuracy on hand written digit dataset
MNIST. Table \ref{tab:cifar10_acc} shows the comparison of classification
accuracy on object recognition dataset CIFAR10.

\begin{table}
  \caption{State of the art SNN accuracies on MNIST dataset. STDP stands for "Spike-Timing-Dependent-Plasticity".}
  \label{tab:mnist_acc}
  \centering
  \begin{tabular}{lll}
    \toprule
    Model & Method & MNIST accuracy \\ 
    \midrule
    Mozafari et al. \cite{Mozafari2018BioInspired} & reward-modulated STDP & 97.2 \\
    Tavanaei et al. \cite{mnist_tavanaei2018} & STDP + gradient descent & 98.6 \\
    Lee et al. \cite{Lee2020} & backpropagation & 99.59 \\
    \textbf{This work} &  \textbf{backpropagation} & \textbf{99.40} \\ 
    \bottomrule
  \end{tabular}
\end{table}

For MNIST dataset we have used a simple convolutional SNN consisting of two
convolutional layers (32 and 64 channels, $5\times 5$ kernels) that are
followed by $2\times 2$ average pooling. After the convolutional part there are
two fully connected layers, the first of which has size 1024. We use dropout
with probability 0.5 before the last layer.  Note, that for such a shallow
network we did not employ batch normalization and used spiking neurons with
soft reset. For evaluation 20 time steps were used. The achieved accuracy is
shown in table~\ref{tab:mnist_acc}.

\begin{table}
  \caption{State of the art SNN accuracies on CIFAR10 dataset.}
  \label{tab:cifar10_acc}
  \centering
  \begin{tabular}{lll}
    \toprule
    Model & Method & CIFAR10 accuracy \\ 
    \midrule
     Wu et al. \cite{Wu2019} & backpropagation & 90.53 \\
     Lee et al. \cite{Lee2020} & backpropagation & 90.95 \\
     Han et al. \cite{Han2020} & ANN-SNN conversion & 93.63 \\
     Rathi et al. \cite{Rathi2020HybridSNN} & ANN-SNN conversion & 92.94 \\
     \textbf{This work} & \textbf{backpropagation} & \textbf{90.20} \\ 
    \bottomrule
  \end{tabular}
\end{table}

For CIFAR10 dataset we used ResNet11 SNN, similar as in \cite{Lee2020} but with
added batch normalization. For evaluation 20 time steps were used. The achieved
accuracy is shown in table~\ref{tab:cifar10_acc}.

\subsection{Scaling up to deeper SNN's}
\label{sec:scaling_up}

In order to determine how training of SNN's with surrogate gradient scales with
number of layers in the network we have trained 5 simple convolutional
classifier architectures with 3, 4, 5, 9 and 13 layers.  The first three
networks have 2, 3 and 4 convolutional layers with stride = 2.  The last two
networks have 4 blocks with 2 and 3 convolutional layers repeated of which only
the first in the block has stride = 2.  The number of channels after the first
convolution is 32 and is subsequently doubled after every layer with stride =
2.  In all of the networks after all of the convolutional layers there is the
last single fully-connected layer mapping the last feature map to the class
confidence scores.  We trained these networks on CIFAR100 object recognition
dataset.  For every architecture we trained 3 variants: ANN, SNN with a fixed
$\gamma=100$ (see eq. \ref{eq:sur-grad-f}, the value was taken from associated
code of \cite{Neftci2019}) and SNN with fine-tuned gamma.  For ANN variants the
ReLU activation was used after every convolutional layer and for SNN's the
simple integrate and fire neurons were used as described in section
\ref{sec:IF_neurons}.

\begin{figure}[tbp]
  \centering
  \includegraphics[width=\linewidth]{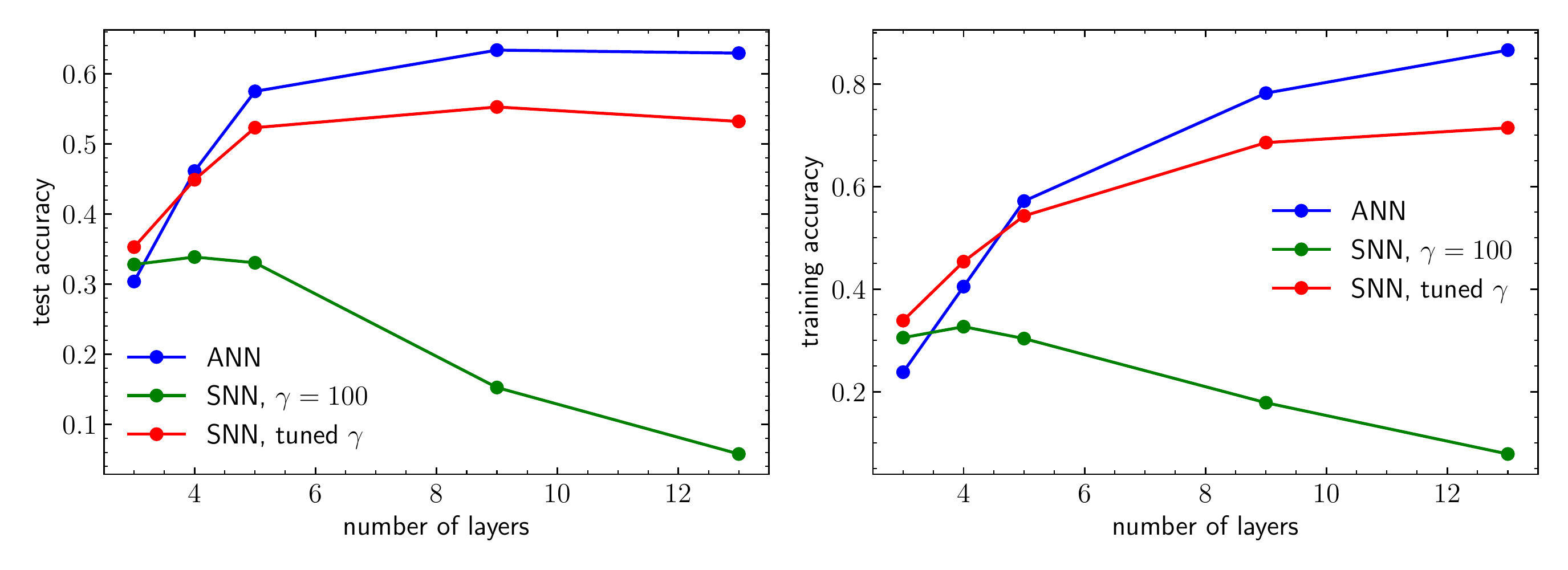}
  \caption{The dependence of final test (left) and training (right) accuracies
  on the number of layers in the network for ANN (blue), SNN with fixed
$\gamma=100$ (green) and SNN with fine-tuned gamma (red).}
  \label{fig:acc_scaling_up}
\end{figure}

The main results of this experiment can be seen in fig.
\ref{fig:acc_scaling_up} where the dependencies of the final testing and
training accuracies on number of layers in these networks are shown. It is
clear that without fine-tuned $\gamma$ only shallow networks can be effectively
trained as already at 5 layers both testing and training accuracies are
starting to decline with increasing number of layers.

It is interesting that with 3 and 4 layers SNNs achieve higher test accuracy
than analogous ANNs (although higher test accuracy is achieved only with 3
layers). This is evidence that SNNs might have higher expressive capacity than
ANNs due to complex time encoding capabilities. However with increasing number
of layers the ANN accuracy grows more quickly and already at 5 layers overcomes
the SNN accuracy. The worse scaling of SNN performance might be explained by
the fact that the surrogate gradient inaccuracy grows larger with increasing
number of layers as described in more detail in section \ref{sec:discussion}.

\begin{table}
  \caption{CIFAR100 accuracy comparison between our ResNet50 SNN and networks obtained by ANN-SNN conversion.}
  \label{tab:cifar100_acc}
  \centering
  \begin{tabular}{lll}
    \toprule
    Model & Method & CIFAR100 accuracy \\ 
    \midrule
     Han et al. \cite{Han2020} & VGG16, ANN-SNN conversion & 70.93 \\
     Rathi et al. \cite{Rathi2020HybridSNN} & VGG11, ANN-SNN conversion & 70.94 \\
     \textbf{This work} & \textbf{Resnet50, backpropagation} & \textbf{58.5} \\
    \bottomrule
  \end{tabular}
\end{table}

To demonstrate training on more complex datasets with very deep network we also
trained ResNet50 SNN on CIFAR100 and a subset of ImageNet \cite{imagenette}. In
the latter case we used images rescaled to $128\times 128$ pixels. For
evaluation 40 and 20 time steps were used respectively. The model achieved
81.2\% accuracy on ImageNet subset and 58.5 \% on CIFAR100. The result of
training on CIFAR100 is shown in table~\ref{tab:cifar10_acc}

\subsection{Accuracy dependence on inference time steps}
\label{sec:time_steps}

\begin{figure}[tbp]
  \centering
  \includegraphics[width=0.6\linewidth]{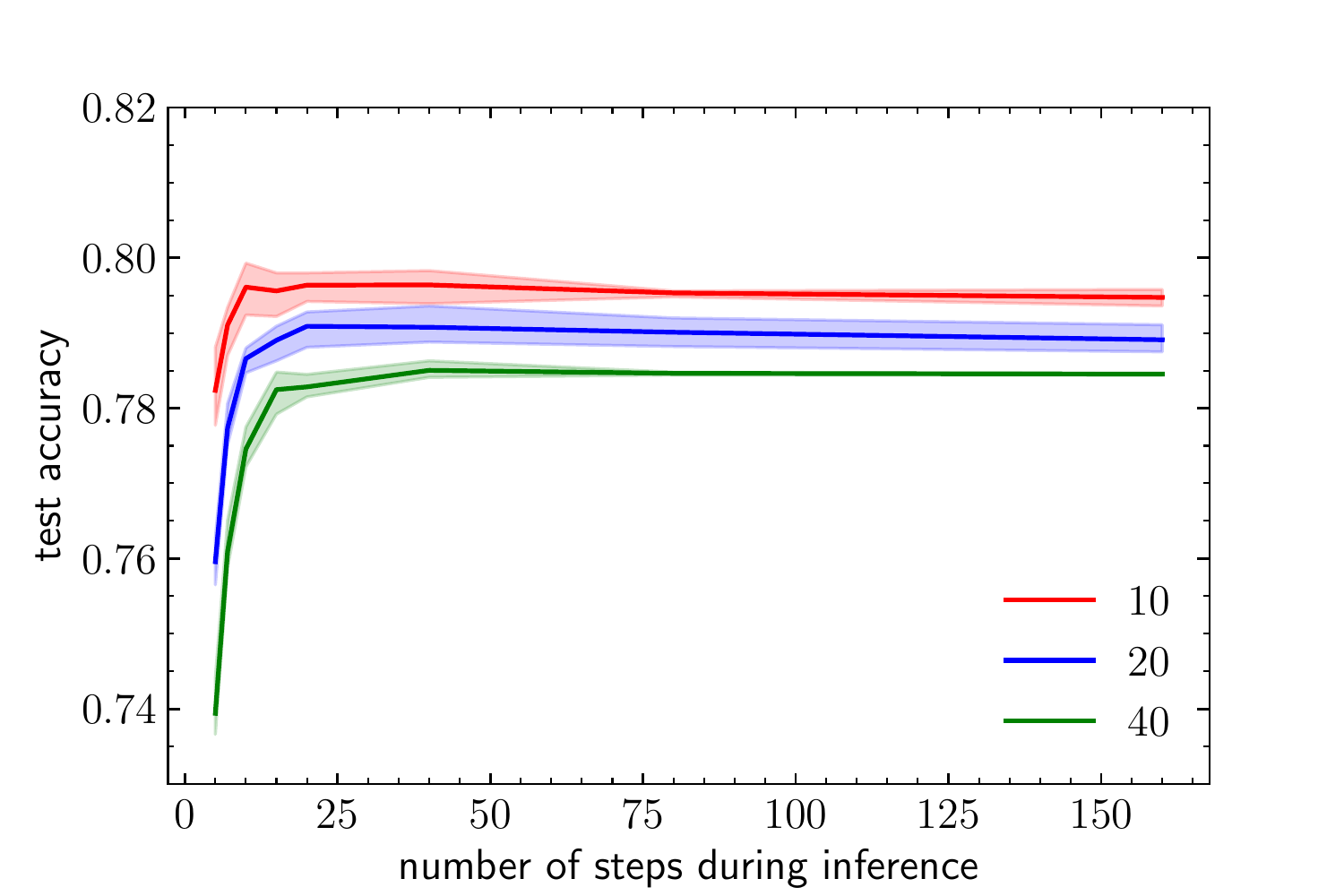}
  \caption{Test accuracy dependence on number of time steps during inference
  for three different numbers of time steps during training: 10 (red), 20
  (blue) and 40 (green). The lines denote the average between the networks with
  different random initializations. The colored areas around the lines denote the
  intervals between minimum and maximum values.}
  \label{fig:acc_vs_time_steps}
\end{figure}

To investigate how our SNN performance varies with different number of time
steps used to simulate the network we trained a small SNN using 10, 20 and 40
time steps. We used the architecture from section \ref{sec:scaling_up} with 4
convolutional layers. To estimate the stochasticity we trained networks with 3
different random parameter initializations in every case. After training the
evaluation can be done with increased or decreased number of time steps. How
this changes the networks accuracy in cases when it was trained with 10, 20 and
40 time steps can be seen in fig. \ref{fig:acc_vs_time_steps}. Interestingly,
we can see that increasing SNN time steps during inference beyond the number
that it was trained with does not improve the accuracy. This behavior of our
SNNs is very different compared to SNNs obtained by conversion from ANNs where
increasing the number of time steps does increase the accuracy and 1000-2000
steps are needed to reach the accuracy close to the original ANN. This
difference most probably arises because our SNNs learn to exploit more
expressive time encoding in order to make predictions in short time instead of
a more inefficient rate encoding used in converted SNNs. The other surprising
trend that can be seen in figure~\ref{fig:acc_vs_time_steps} is that the maximum
accuracy decreases with increasing number of time steps during training. This
might be related to the problematic SNN scaling demonstrated in section
\ref{sec:scaling_up} because the increase of the number of time steps is
similar to the increase of the network depth when using backpropagation through
time (the unrolled computational graph gets deeper).

\section{Discussion and Conclusions}
\label{sec:discussion}

In this work we showed that it is possible to train deep SNNs with minimal
methodological changes compared to ANNs. Moreover, the described method is
simple to implement with popular deep learning frameworks designed primarily
for ANNs by just implementing a custom gradient function for the spiking
activation and then fine-tuning $\gamma$ as described in section
\ref{sec:tuning_gamma}. This makes it is easy to benefit from the vast
contemporary deep learning infrastructure when working with SNNs.  Also, by
using existing deep learning frameworks it is possible to create architectures
incorporating both ANNs and SNNs into a single network which can be trained
end-to-end. This might prove useful in some applications as SNNs might have
advantages in some tasks compared to ANNs and vice versa.

The biggest shortcoming of our method is that with increasing number of layers
in the network eventually the performance starts to decrease and this prevents
the usage of the deepest architectures used in the literature.  Our hypothesis
is that the difference between the effective loss that is minimized by
surrogate gradient and the loss that we actually want to minimize grows with
increasing number of layers or simulated time steps.  We plan to study this
issue in more detail in the future. Still we think that the identification and
solution of exploding and vanishing gradient problem that we presented here is
a significant step forward towards training SNNs of any depth.

Another issue that we think is worthwhile to investigate in the future is that
effective training of SNNs may require significantly different architectures in
comparison to ANNs. In this work we have used only architectures that were
developed for ANNs. This might at least partially explain why it is so hard to
reach ANN performance with SNNs while training models from scratch. 

%% Conclusions
In summary, the main conclusions of this work are the following:
\textbf{i)} exploding and vanishing problems in deep SNNs can be solved by
tuning surrogate gradient width parameter $\gamma$;
\textbf{ii)} batch normalization of the neuron input currents helps training
SNNs similarly like batch normalization in ANNs;
\textbf{iii)} several orders of magnitude less inference time steps are needed
when training SNNs from scratch (in comparison to SNNs converted from ANNs);
\textbf{iv)} with shallow architectures SNNs can sometimes achieve better
performance than analogous ANNs. This might be important in autonomous IoT and
Edge devices;
\textbf{v)} the performance of SNNs trained with surrogate gradients grows
slower than that of ANNs with increasing layers and even starts to diminish at
some point even if the gradients do not explode or vanish.

\section*{Broader Impact}

Currently SNNs still lag behind ANNs in terms of their accuracy on a wide
variety of deep learning tasks. The greatest obstacle for bridging this gap is
training deeper SNNs, and to date training networks deeper than a few layers
remained a challenge. ANN-to-SNN conversion methods, on the other hand require
a separate ANN pretraining step on GPUs and thus are not compatible with
on-device, on-line learning. Thus, effective training methods for SNNs could
enable a wider adoption of energy-efficient neuromorphic hardware in
autonomous, IoT and Edge devices. Also, this kind of hardware might be
especially relevant for applications in space (e.g. nanosatellites).

While our results are promising in terms of supervised training of deep SNNs
approaching that of ANNs, a number of questions remain to be investigated. A
negative outcome of our research would potentially signal a failure of SNNs to
match ANN learning capabilities, and so could result in limited applicability
of deep learning advances for neuromorphic technology.

From ethical standpoint we do not see any special consequences apart from those
associated with general technological progress. 

%\bibliographystyle{plainnat}
%\bibliography{literature}

\end{document}